\documentclass[final]{article}

\usepackage[final]{AdvML_Frontiers_2024}

\usepackage[utf8]{inputenc} 
\usepackage[T1]{fontenc}    
\usepackage{hyperref}       
\usepackage{url}            
\usepackage{booktabs}       
\usepackage{amsfonts}       
\usepackage{nicefrac}       
\usepackage{microtype}      
\usepackage{xcolor}         
\usepackage{natbib}
\usepackage{graphicx}
\graphicspath{{figures/}}

\title{Advancing NLP Security by Leveraging LLMs as Adversarial Engines}

\author{
  Sudarshan Srinivasan\\
  Center for AI Security Research\\
  Oak Ridge National Labratory\\
  Oak Rige, TN 37831\\
  \texttt{srinivasans@ornl.gov}\\
  \And
  Maria Mahbub \\
  Center for AI Security Research\\
  Oak Ridge National Labratory\\
  Oak Rige, TN 37831\\
  \texttt{mahbubm@ornl.gov}\\
  \AND
  Amir Sadovnik \\
  Center for AI Security Research\\
  Oak Ridge National Labratory\\
  Oak Rige, TN 37831\\
  \texttt{sadovnika@ornl.gov}\\
}

\begin{document}

\maketitle

\begin{abstract}
This position paper proposes a novel approach to advancing NLP security by leveraging Large Language Models (LLMs) as engines for generating diverse adversarial attacks. Building upon recent work demonstrating LLMs' effectiveness in creating word-level adversarial examples, we argue for expanding this concept to encompass a broader range of attack types, including adversarial patches, universal perturbations, and targeted attacks. We posit that LLMs' sophisticated language understanding and generation capabilities can produce more effective, semantically coherent, and human-like adversarial examples across various domains and classifier architectures. This paradigm shift in adversarial NLP has far-reaching implications, potentially enhancing model robustness, uncovering new vulnerabilities, and driving innovation in defense mechanisms. By exploring this new frontier, we aim to contribute to the development of more secure, reliable, and trustworthy NLP systems for critical applications.
\end{abstract}

\section{Introduction}

Natural Language Processing (NLP) has been revolutionized by transformer-based \cite{vaswani2017attention} classification models, achieving remarkable success across various domains. These models have become integral to many critical applications, from healthcare to cybersecurity \cite{mahbub2022unstructured, rahali2021malbert, angelis2023energformer}. However, despite their capabilities, these systems remain vulnerable to adversarial attacks \cite{zhang2020adversarial, qiu2022adversarial, goyal2023survey, baniecki2024adversarial}, posing significant risks to their reliability and trustworthiness in crucial sectors.

In this position paper, we argue that leveraging Large Language Models (LLMs) for generating adversarial attacks represents a paradigm shift in NLP security, offering unprecedented opportunities for both attack sophistication and defense enhancement. Recent work has demonstrated the effectiveness of using LLMs for generating valid and natural adversarial examples \cite{wang2024generating}, and we posit that this approach could be extended to address the limitations of current adversarial attack methods, which often produce detectable or semantically incoherent text \cite{jin2020bert, ebrahimi2018hotflip, li2021backdoor}, across various types of attacks including adversarial patches, universal perturbations, and targeted attacks.

LLMs, renowned for their ability to understand and generate human-like text across diverse contexts \cite{minaee2024large}, present a unique opportunity to create adversarial examples that are not only effective at deceiving target classifiers but also indistinguishable from human-written text. This capability could fundamentally change how we approach both the creation of adversarial attacks and the development of robust defenses in NLP. It's crucial to note that we are proposing to use LLMs as tools to generate adversarial patches and not as targets of adversarial attacks.

This proposed approach represents a significant departure from the traditional methods of adversarial attack generation in NLP. By harnessing the sophisticated language understanding and generation capabilities of LLMs, we envision a future where adversarial patches are not just noise in the system, but coherent, context-aware modifications that challenge our very conception of text security. This shift could lead to more robust NLP systems capable of surviving increasingly sophisticated attacks, while also raising new challenges in distinguishing between genuine and adversarial inputs.

However, this novel approach raises important questions: How do we redefine the boundaries between benign and malicious text across different attack types? What are the ethical implications of creating more sophisticated adversarial attacks? How might this approach reshape our understanding of AI security and robustness?

By exploring the potential of LLM-powered adversarial attack generation, we aim to spark discussion on the future of NLP security and the development of more robust AI systems. This paper examines the current challenges in adversarial NLP, presents our position on the transformative potential of LLM-generated adversarial attacks, and discusses the broader implications and future directions of this approach across various attack types.

\section{Current Challenges and Opportunities in NLP Security}

The landscape of NLP security is rapidly evolving, presenting both significant challenges and exciting opportunities. Current adversarial attacks on transformer classifiers encompass a range of techniques, from simple word replacements to more complex perturbations \cite{jin2020bert, ebrahimi2018hotflip, li2021backdoor}. While these methods have shown some success, they face substantial limitations that hinder their effectiveness and applicability in real-world scenarios.

One of the primary challenges across various attack types is the lack of semantic coherence in generated adversarial examples. Many existing techniques produce text that, while successful in fooling models, appears nonsensical or out of context to human readers. This detectability issue severely limits the practical applicability of these attacks, especially in domains where human oversight is common. Additionally, current methods often struggle to maintain the original intent or style of the text while introducing adversarial elements. This is particularly challenging for attacks that aim to be stealthy or preserve specific semantic properties of the original text.

Another crucial limitation is the transferability of adversarial examples. Attacks generated for one model often fail to transfer effectively to other models or domains, restricting their broader impact on NLP security research. This lack of generalizability hampers our ability to develop comprehensive defense strategies against diverse and evolving threats.

However, these challenges also present opportunities for innovation. The emergence of Large Language Models (LLMs) offers a promising avenue for addressing these limitations \cite{wang2024generating}. LLMs have demonstrated remarkable capabilities in understanding and generating human-like text across diverse contexts \cite{minaee2024large}. Their ability to capture long-range dependencies and understand complex language patterns positions them as potential game-changers in the field of adversarial NLP.

We posit that leveraging LLMs for adversarial patch generation could overcome many of the current limitations:
\begin{itemize}
    \item LLMs could generate adversarial examples that maintain contextual relevance and semantic consistency with the original input, regardless of the specific attack type.
    \item Human-like Text: The sophisticated language generation capabilities of LLMs could produce adversarial examples that are indistinguishable from human-written content, enhancing the stealthiness of attacks.
    \item Cross-domain Applicability: Pre-trained on vast amounts of data from various domains, LLMs could potentially generate adversarial examples that are effective across multiple domains and classifier architectures.
    \item Adaptability: The few-shot learning capabilities of many LLMs suggest they could quickly adapt to new tasks or domains with minimal fine-tuning, allowing for the generation of diverse attack types.
    \item Intent Preservation: LLMs' understanding of context and semantics could enable the generation of adversarial examples that preserve the original intent of the text while still fooling classifiers.
\end{itemize}

This novel approach of using LLMs as adversarial engines represents a paradigm shift in how we approach both the creation of adversarial attacks and the development of robust defenses in NLP. By exploring this new paradigm across various attack types, we aim to advance the field of NLP security, potentially leading to more robust and reliable AI systems across various critical applications.

\section{LLMs as Engines for Diverse Adversarial Attacks in NLP}
Recent work by \cite{wang2024generating} has demonstrated the effectiveness of using Large Language Models (LLMs) for generating valid and natural adversarial examples through word-level substitutions. We propose to expand on this foundation, leveraging LLMs as powerful engines for generating a wide range of adversarial attacks in NLP.

Our approach goes beyond word-level modifications to encompass various types of adversarial attacks, including but not limited to:
\begin{itemize}
    \item Adversarial patches: LLMs can generate contextually relevant text snippets that, when inserted into benign inputs, cause misclassification.
    \item Universal perturbations: Utilizing LLMs to create text perturbations that are effective across multiple inputs and potentially multiple target models.
    \item Targeted attacks: Employing LLMs to craft adversarial examples aimed at specific misclassifications, leveraging their deep understanding of language and context.
    \item Transferable attacks: Exploiting LLMs' broad knowledge to generate adversarial examples that are effective across different model architectures and domains.
\end{itemize}

\begin{figure}[htbp]
    \centering
    \includegraphics[width=0.8\textwidth]{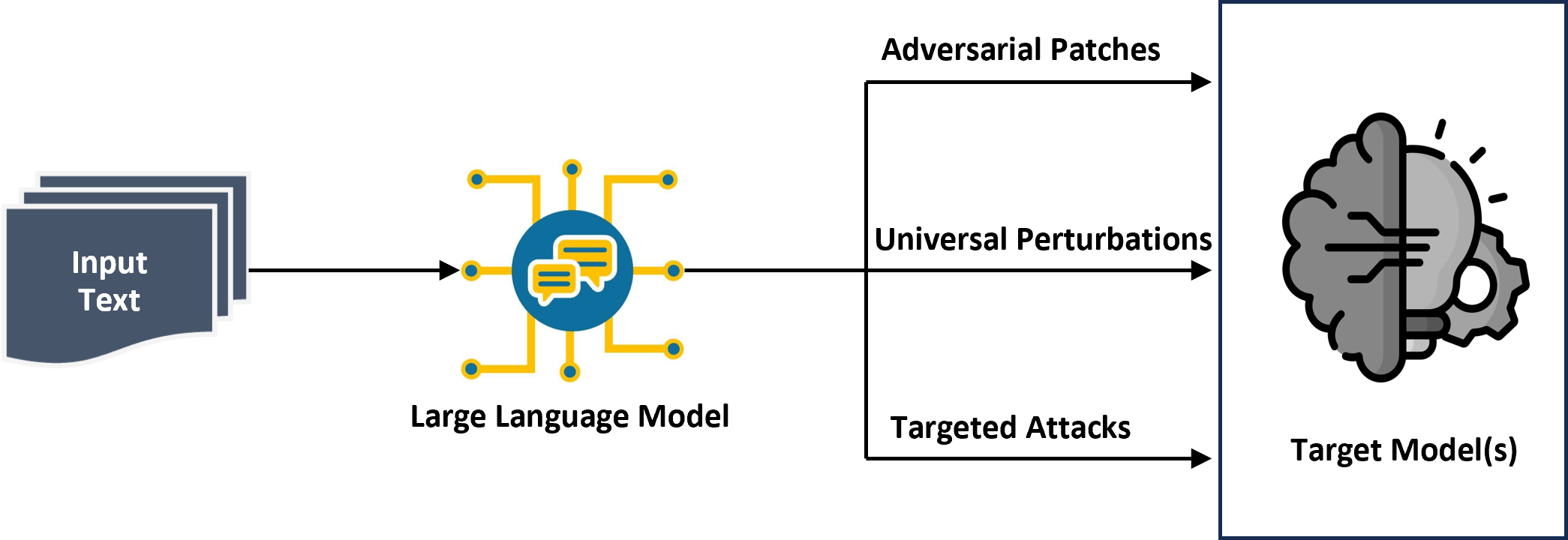}
    \caption{Conceptual Framework of LLM-Powered Adversarial Attack Generation for NLP System}
    \label{fig:framework}
\end{figure}

We propose a novel paradigm for generating adversarial patches in NLP using Large Language Models (LLMs) shown in figure \ref{fig:framework}. This approach represents a fundamental shift in how we conceptualize and create adversarial examples for text data. Unlike traditional methods that rely on simple word replacements or character-level modifications, our proposed approach leverages the contextual understanding of LLMs. This allows for the generation of adversarial examples that seamlessly integrate with the surrounding text, making them significantly more challenging to detect. We envision a process where LLMs are fine-tuned or prompted to generate adversarial examples based on specific attack goals and constraints. This could involve iterative refinement, where the LLM generates candidates, receives feedback on their effectiveness, and improves its outputs accordingly.

By expanding the use of LLMs beyond word-level substitutions to a comprehensive adversarial engine, we aim to push the boundaries of what's possible in adversarial NLP. In addition to sophisticated attacks on transformer-based models, this approach has the potential to uncover previously unknown vulnerabilities in NLP systems. It enables us to develop more comprehensive and realistic datasets for adversarial training.

However, this paradigm also raises important questions and challenges. How do we ensure ethical use of such powerful adversarial generation capabilities? What new defense mechanisms will be needed to counter these more sophisticated attacks? How might this approach influence the development of future NLP models and architectures?

By exploring these questions and pushing the boundaries of adversarial NLP, we believe this new paradigm has the potential to significantly advance the field of AI security, leading to more robust, reliable, and trustworthy NLP systems.

\section{Implications and Future Directions}
The proposed approach of using LLMs as engines for diverse adversarial attacks in NLP has far-reaching implications for both offensive and defensive aspects of AI security. One of the most significant implications is the potential to enhance the robustness of transformer-based classifiers through advanced adversarial training. By generating large-scale datasets of sophisticated, human-like adversarial examples across various attack types, we can train classifiers to be more resilient against a wide range of potential attacks \cite{yoo2021towards,yang2024fast}. This could lead to the development of more secure and reliable AI systems, particularly in critical applications such as healthcare, cybersecurity, and energy infrastructure \cite{patwardhan2023transformers}.

The ability to generate human-like adversarial examples across different attack types raises important questions about the nature of AI vulnerabilities. As these examples become increasingly indistinguishable from genuine human input, it may necessitate a reevaluation of what constitutes an adversarial example and how we define model robustness \cite{yuan2021current}. This could lead to new theoretical frameworks for understanding and quantifying the security of NLP systems.

From an offensive security perspective, the proposed approach could potentially reveal previously unknown vulnerabilities in existing NLP systems. By systematically exploring the space of possible adversarial attacks using LLMs, we may uncover new attack vectors that current defense mechanisms are ill-equipped to handle \cite{li2021exploring}. This knowledge, while potentially concerning, is crucial for developing more comprehensive defense strategies.

The use of LLMs in generating diverse adversarial attacks opens up interesting research directions in the field of AI alignment. As we leverage one AI system (the LLM) to generate attacks against another (the target classifier), we may gain new insights into the interplay between different AI architectures and the nature of machine-to-machine interactions in adversarial settings \cite{ji2023ai}.

Looking to the future, this research could pave the way for more sophisticated, context-aware defense mechanisms in NLP. As adversarial attacks become more advanced, so too must our methods for detecting and mitigating their effects. This might involve developing new techniques for distinguishing between genuine and artificially generated text, or creating adaptive defense systems that can recognize and neutralize emerging attack patterns in real-time \cite{goyal2023survey,minh2023advanced,qiu2022adversarial}.

The ethical implications of this research warrant careful consideration and further study. The ability to generate highly convincing adversarial examples across various attack types raises questions about potential misuse, such as in the creation of sophisticated disinformation campaigns \cite{garg2023comprehensive}. Future work should focus on developing ethical guidelines and safeguards for the responsible development and use of these technologies.

While the proposed approach shows promise, it's important to acknowledge potential limitations and risks. The computational cost of fine-tuning and using large language models for adversarial attacks may be prohibitive for some applications. There is also a risk of overfitting, where LLM-generated examples might become too specific to certain models or datasets, limiting their generalizability. If this approach proves less effective than anticipated, alternative directions could include exploring hybrid approaches that combine traditional adversarial techniques with LLM capabilities, focusing on improving the interpretability of NLP models, developing more sophisticated ensemble methods for robust NLP systems, or investigating the use of formal verification techniques in NLP security.

We believe that the proposed approach of using LLMs for generating diverse adversarial attacks represents a significant step forward in the field of NLP security. It not only offers new tools for testing and improving the robustness of AI systems but also opens up exciting new avenues for research in adversarial machine learning, AI alignment, and ethical AI development.

\section{Conclusion}
In this position paper, we have presented a novel perspective on the future of adversarial machine learning in NLP, proposing the use of Large Language Models as powerful engines for generating diverse adversarial attacks. This approach represents a significant advancement from recent work that has demonstrated the effectiveness of LLMs in generating word-level adversarial examples \cite{wang2024generating}.

We argue that leveraging LLMs for adversarial attack generation has the potential to:
\begin{itemize}
    \item Create more effective and human-like adversarial examples across various attack types, including adversarial patches, universal perturbations, and targeted attacks.
    \item Uncover new vulnerabilities in existing NLP systems, pushing the boundaries of what we consider ``secure'' in NLP.
    \item Enhance the robustness of AI models through advanced adversarial training using more sophisticated and diverse adversarial examples.
    \item Drive innovation in defense mechanisms to counter these more advanced attacks.
    \end{itemize}
    
However, this approach also raises important ethical considerations and challenges that the research community must address. As we move forward, it will be crucial to develop this technology responsibly, with a focus on enhancing the overall security and reliability of NLP systems.

The interdisciplinary nature of this research opens up exciting possibilities for collaboration across various fields, including machine learning, linguistics, cybersecurity, and ethics. These collaborations will be essential in addressing the complex challenges that arise from more sophisticated adversarial techniques.

As AI systems continue to play an increasingly critical role in our society, ensuring their security and reliability becomes ever more important. We believe that this work will contribute to the ongoing effort to create more robust, trustworthy AI systems that can withstand sophisticated adversarial attacks while maintaining their performance and utility.

In conclusion, while challenges remain, the potential of LLM-powered adversarial attack generation to revolutionize NLP security is significant. We hope this position paper will spark further discussion and research in this exciting and important area, ultimately leading to more secure and reliable NLP systems across various critical applications.

\section*{Acknowledgements}
This manuscript has been in part co-authored by UT-Battelle, LLC under Contract No. DE-AC05-00OR22725 with the U.S. Department of Energy.

\bibliographystyle{plainnat}
\bibliography{references}

\end{document}